\pdfoutput=1

\documentclass[11pt,table,dvipsnames]{article}

\usepackage[final]{acl}
\usepackage{times}
\usepackage{latexsym}

\usepackage[T1]{fontenc}

\usepackage[utf8]{inputenc}

\usepackage{microtype}

\usepackage{inconsolata}
\usepackage{microtype}
\usepackage{latexsym}
\usepackage{dsfont}
\usepackage{booktabs} 
\usepackage{CJKutf8}
\usepackage{graphicx}
\usepackage{bigstrut}
\usepackage{multirow}
\usepackage{amsmath}
\usepackage{multirow, makecell, caption}
\usepackage{floatrow}
\newfloatcommand{capbtabbox}{table}[][\FBwidth]
\usepackage{xcolor}
\usepackage[normalem]{ulem}
\usepackage{amssymb}
\usepackage{amsfonts}
\usepackage{multicol}
\usepackage{tipa}
\usepackage{amsmath}
\usepackage{bm}
\usepackage[ruled,linesnumbered]{algorithm2e}
\usepackage{tikz}
\usepackage{paralist}
\usepackage{IEEEtrantools}
\usepackage[switch]{lineno}
\usepackage{enumitem}
\usepackage{multirow, makecell, caption}
\usepackage{arydshln}
\usepackage{verbatim}
\usepackage[normalem]{ulem}
\usepackage{amssymb}
\usepackage{amsfonts}
\usepackage{multicol}
\usepackage{amssymb}
\usepackage{pifont}
\usepackage{csquotes}
\usepackage{xspace}
\usepackage{paralist}
\usepackage{mdwlist}
\usepackage{subcaption}
\usepackage{makecell}
\usepackage{array}
\usepackage{bm}
\usepackage{colortbl}
\usepackage{dsfont}
\usepackage{url}
\usepackage{booktabs} 
\usepackage{graphicx}
\usepackage{tabularx}
\usepackage{amsmath}
\usepackage{bbm}
\usepackage{pgfplots}
\usepackage{soul} 
\usepackage{listings}
\usepackage[tableposition=top]{caption}
\usepackage{color,xcolor} 

\newcommand{\ie}{\textit{i.e.}\xspace}
\newcommand{\eg}{\textit{e.g.}\xspace}

\newcommand{\task}{\textit{Motivation Recognition}\xspace}
\newcommand{\method}{\texttt{SuperSummary}\xspace}
\newcommand{\dataset}{\textsc{CroSS}\xspace}
\newcommand{\gptfour}{GPT-4\xspace}
\newcommand{\llama}{Llama-3-70B\xspace}

\makeatletter
\def\adl@drawiv#1#2#3{%
        \hskip.5\tabcolsep
        \xleaders#3{#2.5\@tempdimb #1{1}#2.5\@tempdimb}%
                #2\z@ plus1fil minus1fil\relax
        \hskip.5\tabcolsep}
\newcommand{\cdashlinelr}[1]{%
  \noalign{\vskip\aboverulesep
           \global\let\@dashdrawstore\adl@draw
           \global\let\adl@draw\adl@drawiv}
  \cdashline{#1}
  \noalign{\global\let\adl@draw\@dashdrawstore
           \vskip\belowrulesep}}
\makeatother

\title{Evaluating Character Understanding of Large Language Models via Character Profiling from Fictional Works}

\author{Xinfeng Yuan\textsuperscript{\rm $\heartsuit$},
 Siyu Yuan\textsuperscript{\rm $\heartsuit$}\thanks{~~Equal contribution.},
 Yuhan Cui\textsuperscript{\rm $\spadesuit$}\footnotemark[1],
 Tianhe Lin\textsuperscript{\rm $\heartsuit$},\\
 \textbf{Xintao Wang}\textsuperscript{\rm $\spadesuit$},
 \textbf{Rui Xu}\textsuperscript{\rm $\spadesuit$},
 \textbf{Jiangjie Chen}\textsuperscript{\rm $\spadesuit$},
 \textbf{Deqing Yang}\textsuperscript{\rm $\heartsuit$}\thanks{~~Corresponding author.}\\
\textsuperscript{\rm $\heartsuit$}School of Data Science, Fudan University\\
\textsuperscript{\rm $\spadesuit$}Shanghai Key Laboratory of Data Science, School of Computer Science, Fudan University\\
\texttt{\{xfyuan23,syyuan21,xtwang21,ruixu21\}@m.fudan.edu.cn}\\
\texttt{\{yhcui20,thlin20,jjchen19,yangdeqing\}@fudan.edu.cn}}
\begin{document}

\maketitle

\begin{abstract}
Large language models (LLMs) have demonstrated impressive performance and spurred numerous AI applications, in which role-playing agents (RPAs) are particularly popular, especially for fictional characters. 
The prerequisite for these RPAs lies in the capability of LLMs to understand characters from fictional works. 
Previous efforts have evaluated this capability via basic classification tasks or characteristic imitation, failing to capture the nuanced character understanding with LLMs. 
In this paper, we propose evaluating LLMs' character understanding capability via the character profiling task, \ie, summarizing character profiles from corresponding materials, a widely adopted yet understudied practice for RPA development. 
Specifically, we construct the \dataset dataset from literature experts and assess the generated profiles by comparing them with ground truth references and evaluating their applicability in downstream tasks.
Our experiments, which cover various summarization methods and LLMs, have yielded promising results.
These results strongly validate the character understanding capability of LLMs.
Resources are available at \url{https://github.com/Joanna0123/character_profiling}.

\end{abstract}

\section{Introduction}
\label{sec:intro}

\begin{figure*}[t]
    \centering
    \includegraphics[width=0.85\linewidth]{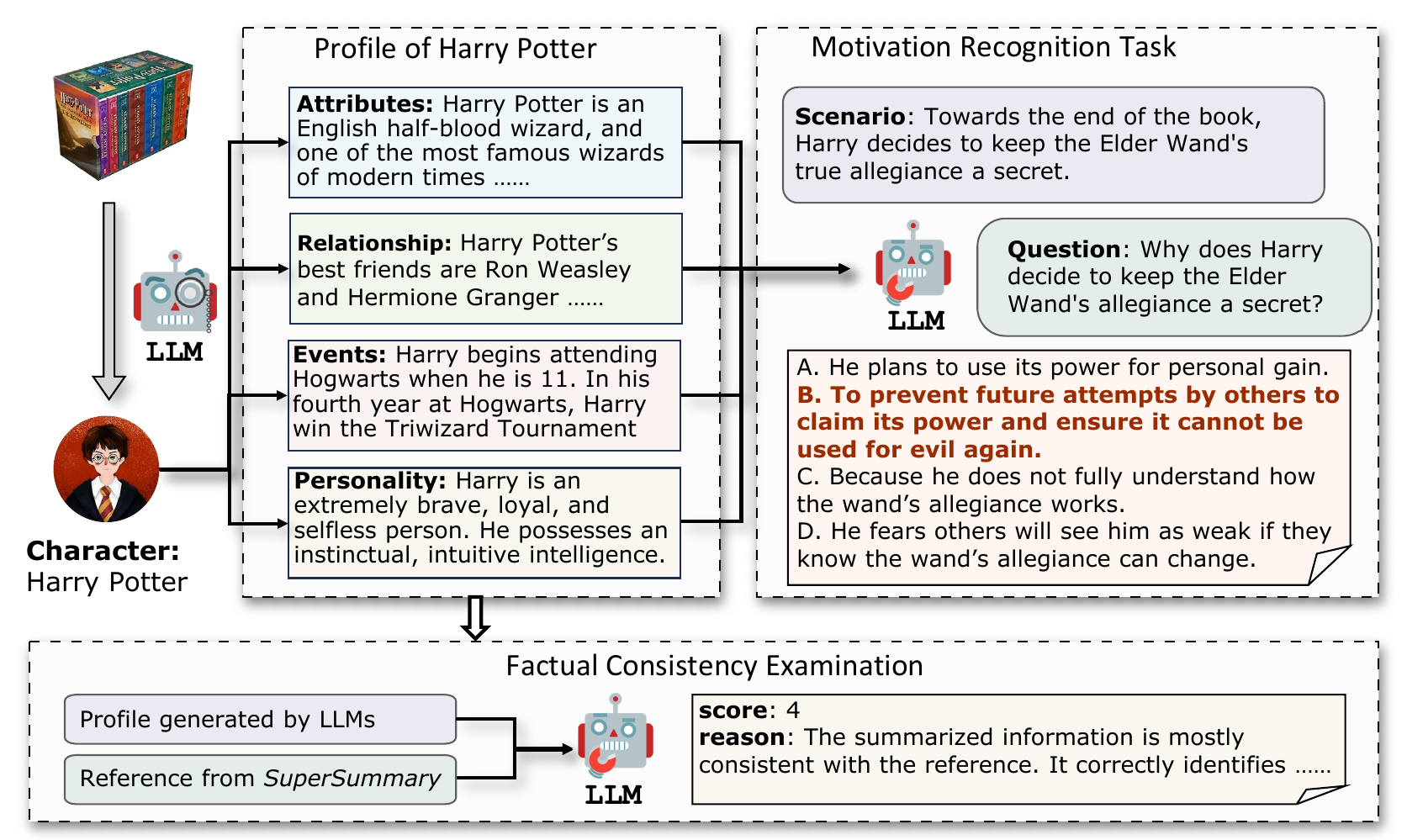}
    \vspace{-0.2cm}
    \caption{An overview of character profiling with LLMs and the two evaluation tasks we proposed, including factual consistency examination and motivation recognition.}
    \label{fig:front}
\end{figure*}

The recent progress in large language models (LLMs)~\cite{achiam2023gpt,anthropic2024claude3} has catalyzed numerous AI applications, among which role-playing agents (RPAs) have attracted a wide range of audiences.
RPAs are interactive AI systems that simulate various personas for applications, including chatbots of fictional characters~\cite{wang2023rolellm}, AI none player characters in video games~\cite{wang2023voyager}, and digital replicas of real humans~\cite{gao2023livechat}. 
In practice, LLMs are generally prompted with character profiles to role-play fictional characters~\citep{Wang2023InCharacterEP, zhao2023narrativeplay}, and these profiles are typically generated through the automatic summarization of corresponding literature using advanced LLMs~\cite{wang2023rolellm,li2023chatharuhi}. 

Previous efforts have studied LLMs' capabilities of understanding 
characters from fictional works. 
The research on character understanding mainly concentrates on basic classification tasks, such as character prediction~\cite{brahman2021let, yu2022few,li2023multi} and personality prediction~\cite{yu2023personality}, which aims at recognizing characters or predicting their traits from given contexts correspondingly. 
Recently, the research focus has shifted to character role-playing, primarily focusing on the imitation of characteristics such as knowledge~\cite{Tang2024EnhancingRS,Shen2023RoleEvalAB} and linguistic style~\cite{zhou2023characterglm,wang2023rolellm}.
Hence, these tasks fail to capture the nuanced character understanding of LLMs.

In this paper, we systematically evaluate LLMs' capability on the \textbf{character profiling} task, \ie, summarizing profiles for characters from fictional works. 
For research, character profiling is indeed the first task to explore the depth of LLMs' character understanding via generation.
This is more challenging than previous classification tasks, contributing to a more nuanced comprehension of how LLMs understand the character.  
In practice, the character profiles generated by LLMs have been widely adopted for RPA development~\citep{wang2023rolellm,li2023chatharuhi, xu2024character}, and have the potential to facilitate human understanding of characters, but their effectiveness remains significantly understudied.
Our work in this paper aims to evaluate LLMs' performance on character profiling, of which the challenges mainly include the absence of high-quality datasets and evaluation protocols. 

To address these challenges, we construct the \dataset (\textbf{C}haracter P\textbf{ro}files from \textit{\textbf{S}uper\textbf{S}ummary}) dataset for character profiling, and propose two tasks to evaluate the generated profiles. 
The \dataset dataset is sourced from \method~\footnote{\url{https://www.supersummary.com}}, a platform providing summaries for books and characters contributed by literature experts.
Our evaluation distinguishes four essential dimensions for character profiles: attributes, relationships, events, and personality.
We parse the character profiles from \method into these dimensions by \gptfour, as the ground truth references.
Then, the generated profiles are evaluated in either an intrinsic or extrinsic way.
The intrinsic evaluation directly employs \llama\cite{llama3modelcard} to compare the generated profiles with the references. 
For extrinsic evaluation, we propose the \task task and 
measure whether the generated profiles can support LLMs in this task, \ie, identifying the motivations behind characters' decision-making.

Our experiments cover various summarization methods, including \emph{Hierarchical Merging}, \emph{Incremental Updating}, and \emph{Summarizing in One Go}, implemented on numerous LLMs.  
The results reveal that character profiles generated by LLMs are satisfactory but leave space for further improvement. 
This suggests the potential information loss in RPAs built on these profiles.
Additionally, the results of \task demonstrate the importance of each of the four dimensions for character profiles.

Our contributions are summarized as follows:
\begin{inparaenum}[\it 1)]
\item We present the first work to evaluate LLMs' capability of character profiling and propose an evaluation framework with detailed dimensions, tasks, and metrics.
\item We introduce \dataset, a high-quality dataset valuable for character profiling tasks, which is sourced from literature experts.  
\item We conduct extensive experiments with different summarization methods and LLMs, showcasing the promising effectiveness of using LLMs for character profiling.
\end{inparaenum}

\section{Related Work}
\label{sec:related}

\paragraph{Character Role-Playing}
Recent advancements in LLMs have significantly enhanced the capabilities of role-playing agents (RPAs) across various aspects.
Currently, many role-playing tasks require interactive AI systems to act as assigned personas, including celebrities and fictional characters.
In these studies, researchers have utilized various methods to develop RPAs, which can be divided into three categories:
\begin{inparaenum}[\it 1)]
    \item \textit{Manual Construction}~\cite{chen2023large, zhou2023characterglm}, which employs book fans or professional annotators to label information related to characters;
    \item \textit{Online Resource Collection}~\cite{shao2023character, tu2024charactereval}, which collects character profiles from online resources, 
    \eg, Wikipedia~\footnote{\url{https://en.wikipedia.org/wiki/Main_Page}}, and Baidu Baike~\footnote{\url{https://baike.baidu.com/}};
    \item \textit{Automatic Extraction}~\cite{li2023chatharuhi,zhao2023narrativeplay}, which utilizes LLMs to extract character dialogues from origin books or scripts.
\end{inparaenum}
In this paper, we explore the capabilities of LLMs in generating character profiles for RPAs construction.

\paragraph{Motivation Analysis \& Character Understanding}
Motivation is a fundamental concept, which is shaped by personality traits and the immediate surroundings~\cite{young1961motivation,atkinson1964introduction,kleinginna1981categorized}.
In narrative texts, the motivation of a character can reveal their inner traits and their relationship with the external world. Thus, understanding the motivation of characters strongly aligns with the LLMs' ability to comprehend characters.
Previous studies typically propose benchmarks in character identification~\cite{chen2016character, brahman2021let, sang2022tvshowguess, yu2022few}, situated personality prediction~\cite{yu2023personality}, question answering~\cite{kovcisky2018narrativeqa}.
Despite these efforts, prior research has not focused on assessing a character's motivation based on character profiles.
To bridge this gap, we propose the motivation recognition task. This task aims to directly evaluate whether LLMs can grasp a character's essence by identifying the motivations behind each decision within a story.

\section{Character Profiling Framework}
\label{sec:method}
\begin{figure*}[ht]
    \centering
    \begin{subfigure}[b]{0.36\textwidth}
        \centering
        \includegraphics[height=4.5cm, keepaspectratio]{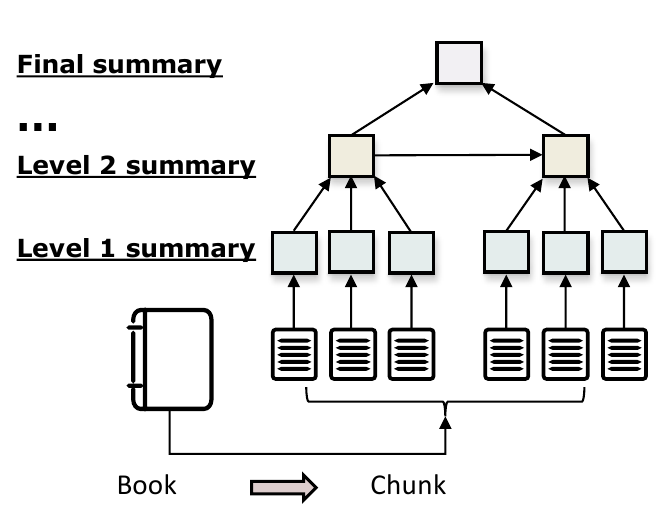}
        \caption{Hierarchical Merging}
        \label{fig:hier}
    \end{subfigure}
    \hfill
    \begin{subfigure}[b]{0.36\textwidth}
        \centering
        \includegraphics[height=4.5cm, keepaspectratio]{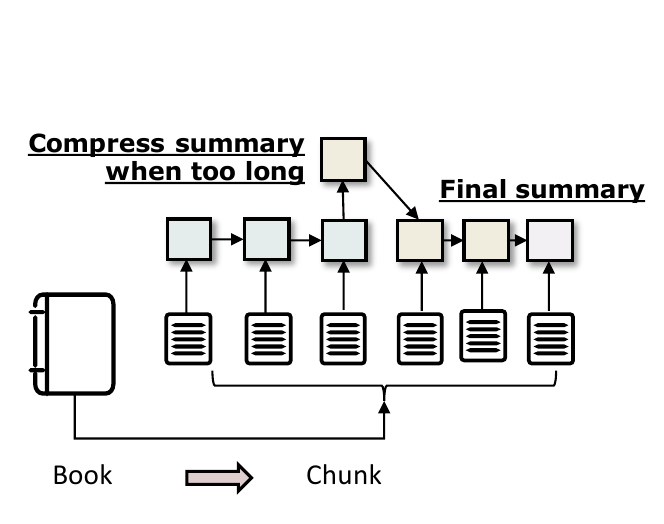}
        \caption{Incremental Updating}
        \label{fig:inc}
    \end{subfigure}
    \hfill
    \begin{subfigure}[b]{0.25\textwidth}
        \centering
        \includegraphics[height=4.5cm, keepaspectratio]{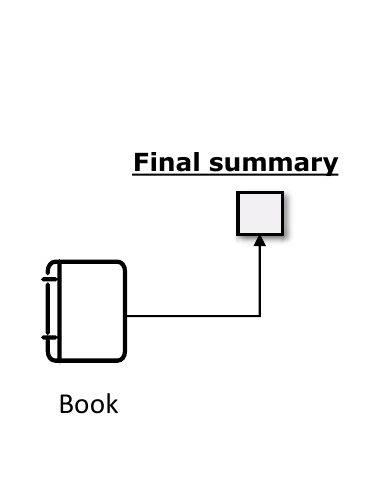}
        \caption{Summarizing in One Go}
        \label{fig:once}
    \end{subfigure}
    \caption{The three methods of long context processing for LLM-based character profiling.}
    \label{fig:method}
\end{figure*}

\subsection{Task Formulation}
Character profiling aims to generate profiles for fictional characters from corresponding literature.
Given the input character name $\mathcal{N}$ and the original content $\mathcal{B}$ of a fictional work, the LLM should output the character profile $\mathcal{P}$ which covers the core information about the character.
Specifically, in this paper, 
$\mathcal{P}= (\mathcal{P}_{attributes}, \mathcal{P}_{relationships}, \mathcal{P}_{events}, \mathcal{P}_{personality})$ is structured in four dimensions, as detailed in Section~\ref{sec:dims}. 
An example of a character profile is presented in Figure~\ref{fig:front}.

\subsection{Character Profile Dimensions}
\label{sec:dims}
For a character, the profile should be highly complex and multi-faceted, embodying diverse information.
Drawing inspiration from previous studies and current developments in persona products~\cite{zhao2023narrativeplay, baichuannpc}, we define four main profile dimensions for LLMs to summarize, which are commonly examined in literary studies~\cite{yu2023personality,Zhao2024LargeLM, Shen2023RoleEvalAB}. Please refer to Appendix \ref{appendix:comparison} for a further comparison.

\paragraph{Attributes}
The basic attributes of a character encompass gender, skills, talents, objectives, and background.

\paragraph{Relationships}
A character's interpersonal relationships are a vital aspect of their profile, which are intimately connected to the character's experiences and personality. 
Moreover, these relationships can serve as a foundation for constructing fictional character relationship diagrams.

\paragraph{Events}
Events cover the experiences that characters have been part of or impacted by, marking a critical profile dimension. 
Due to the complexity of certain narratives, such as alternating timelines and showcasing events from diverse worlds or different perspectives, we require the model to rearrange events and order them chronologically.

\paragraph{Personality}
Personality refers to the lasting set of characteristics and behaviors that form an individual's unique way of adapting to life~\cite{APA_Dictionary}.
While well-rounded characters often exhibit complex personalities, their personalities can be analyzed through their actions, choices, and interactions with others.

\subsection{Summarization Methods}
Book-length texts often comprise over 100,000 tokens, surpassing the context window limitations of many current LLMs.
As a result, the primary framework for long context processing involves segmenting books into manageable segments for LLMs, followed by subsequent comprehensive processing.
As illustrated in Figure~\ref{fig:hier} and Figure~\ref{fig:inc}, we inherit two methods for book summarization~\cite{chang2023booookscore}, \ie, hierarchical merging and incremental updating.
Additionally, for models that can handle long context windows, we explore the method of summarizing in one go, as shown in Figure~\ref{fig:once}.

\paragraph{Hierarchical Merging}

The hierarchical merging approach~\cite{wu2021recursively} employs a simple, zero-shot prompt technique. 
It begins by summarizing information from segments within a book, generating the summaries at level 1. 
Then, several summaries are combined to establish the initial context at level 2. 
Subsequently, it merges the following summaries with context iteratively. 
The merging process continues at the next level until a final summary is generated.

\paragraph{Incremental Updating}

One major issue with hierarchical methods lies in constructing summaries at level 1.
As shown in Figure~\ref{fig:hier}, the provided text only contains novel content from the current segment without any background information from earlier segments. 
Thus, this absence of context may increase the risk of misinterpreting information in later segments.

In response, \citet{chang2023booookscore} introduces incremental updating.
This method leverages background information from the preceding text to enhance summary quality. 
The process of incremental updating consists of three phases:
First, it starts by summarizing the book's opening segment.
Then, this summary is refined and updated by incorporating details from the following segments recursively.
Throughout this process, to ensure conciseness and relevance, the summary is periodically condensed to comply with a set pre-defined maximum length.
By following these steps, the method seeks to promote a more integrated and coherent comprehension of the entire text.

\paragraph{Summarizing in One Go}
Recent developments in LLMs have introduced models capable of processing over 100,000 tokens. 
For example, GPT-4-Turbo~\cite{achiam2023gpt} supports a context window of up to 128,000 tokens. 
This advancement enables us to explore a method for inputting the full content of a book into the model in one step. 
For this investigation, we select books from our dataset that contain fewer than 120,000 tokens.
\begin{table}[t]
  \centering
  \small
    \begin{tabularx}{\linewidth}{X}
    \toprule
    \textbf{Character}: Nora Stephens \\
    \textbf{Character Profile}: \\
    Attributes: Nora Stephens is a tall, thin, ambitious female literary agent with dyed platinum blonde hair, known for her pragmatic approach and dedication to her career. She has a history of troubled$\dots$\\
    
    \textbf{Scenario}: \\
    Nora decides not to pursue a long-distance relationship with Charlie after the summer ends. \\
    \textbf{Question}: Why does Nora make the decision to part ways with Charlie?\\
    \textbf{Options}: \\
    \textcolor{red}{\textit{A. Because she fears long-distance relationships are doomed to fail.}}\\
    B. Because she believes she needs to focus on her personal growth and independence.\\
    C. Because she feels their goals and aspirations are no longer aligned.\\
    D. Because she worries that their frequent arguments are harming her well-being.\\
    \midrule
    \rowcolor[gray]{0.95} \textbf{Model Reasoning Output}\\
    \textbf{Choice}: "A", \\
    \textbf{Reason}: Nora's decision $\dots$ is primarily motivated by her fear that long-distance relationships are doomed to fail. Given her history of being dumped and her protective nature due to her family's past, Nora is likely cautious about entering a relationship that has inherent challenges and uncertainties$\dots$ \\
    \bottomrule
    \end{tabularx}
  \caption{A toy example of MR task. A complete set of data includes character name, character profile, scenario, question, options, correct answer, and reason. The reasoning model is GPT-4-Turbo-0409 \protect\footnotemark[1].}
  \label{tab:case}
\end{table}
\footnotetext[1]{In this paper, unless otherwise specified, we adopt the version of GPT-4-Turbo-0125 throughout.}

\section{Evaluation Protocol}
\label{sec:evaluation}

\subsection{Evaluation Tasks}

\paragraph{Intrinsic Evaluation: Factual Consistency Examination (FCE)}
To generate character profiles from books, we implement the three methods previously described.
Throughout the summarization process, we require the model to produce four distinct sections, each detailing one dimension of a character's profile.
An excellent profile should accurately cover all the important information about the character across these four dimensions. 
Therefore, we evaluate factual consistency by comparing the model-summarized profile with the reference profile.
The metrics for this examination are introduced in Section~\ref{sec:ca_metric}.

\paragraph{Extrinsic Evaluation: Motivation Recognition (MR)}
As shown in Table~\ref{tab:case}, to thoroughly evaluate whether the summarized profiles enhance models' understanding of a character's essence, we introduce a \task task for downstream evaluation. 
This task investigates if the character profiles generated by the model effectively aid in comprehending the characters, particularly in recognizing the motivations behind their decisions.

Given the input $\mathcal{X} = (\mathcal{N}, \mathcal{P}, \mathcal{D}, \mathcal{Q}, \mathcal{A})$, which includes the character name $\mathcal{N}$, the character profile $\mathcal{P}$ defined by four dimensions, the character's decision $\mathcal{D}$, a question $\mathcal{Q}$ about the motivations behind the decision, and a set of potential answer $\mathcal{A} = \{a_i\}_{i=1}^4$ for $\mathcal{Q}$, the LLMs should determine the answer $\mathcal{Y}$ from $\mathcal{A}$ that correctly reflects the character's motivation.
Details of MR dataset construction are provided in Section \ref{sec:mgchoice}.

\subsection{Evaluation Metrics}
\newcolumntype{a}{>{\columncolor{BlueGreen!10}}c}
\newcolumntype{b}{>{\columncolor{brown!10}}r}
\newcolumntype{d}{>{\columncolor{blue!10}}r}
\newcolumntype{q}{>{\columncolor{Green!10}}r}
\begin{table*}[t]
\centering  
\small
\begin{tabular}{llbbbbbd}
\toprule
\multicolumn{1}{c}{\multirow{2}[1]{*}{\makecell{\textbf{Summarization}\\\textbf{Method}}}} &  
\multicolumn{1}{c}{\multirow{2}[1]{*}{\textbf{Summarization Model}}} &
\multicolumn{5}{c}{\textbf{Consistency Score}} & 
\multicolumn{1}{c}{\multirow{2}[1]{*}{\makecell{\textbf{MR}\\\textbf{Acc.}}}}\\

\cmidrule(lr){3-7}
& & \multicolumn{1}{c}\textbf{Attr} & \multicolumn{1}{c}\textbf{Rela} & \multicolumn{1}{c}\textbf{Even} & \multicolumn{1}{c}\textbf{Pers} & \multicolumn{1}{c}\textbf{Avg.}\\
\midrule
\rowcolor[gray]{0.95} \multicolumn{8}{c}{\textit{\dataset (Full dataset)}} \\
\midrule
\addlinespace[0.05cm]
\multirow{9}{*}{\makecell[c]{\textbf{Incremental}\\\textbf{Updating}}} & Mistral-7B-Instruct-v0.2 &  2.75&	2.20	& 1.88	&3.89	&2.68  &48.31\\
 & Mixtral-8x7B-MoE &2.75	&2.58	& 2.28	&4.02	&2.91	 & 52.13\\
 & vicuna-7b-v1.5-16k &2.44	&1.72	&1.45	&3.17	&2.20	 & 42.70\\
 & vicuna-13b-v1.5-16k &2.79	&2.22	&1.76	&3.56	& 2.58 & 46.29\\
 & Qwen1.5-7B-Chat & 2.35 &1.98	&1.58	&3.75	&2.42 & 44.49 \\
 & Qwen1.5-14B-Chat& 2.39	&2.18	&1.41	&3.74	&2.43 & 47.42\\
 & Qwen1.5-72B-Chat &3.33	&2.71	&2.45	&4.08	&3.14 & 52.36\\
 & GPT-3.5-Turbo &3.49	&2.57	&1.95	&3.95	&2.99 & 49.44\\
 & GPT-4-Turbo & \underline{3.72}	&\underline{3.24}	&\textbf{3.58}	&3.87	&\underline{3.60} & \textbf{57.75}\\
\cdashlinelr{1-8}
\multirow{9}{*}{\makecell[c]{\textbf{Hierarchical}\\\textbf{Merging}}} & Mistral-7B-Instruct-v0.2 &3.07	&2.20	&1.98	&3.83	&2.77 & 50.56\\
 & Mixtral-8x7B-MoE &3.17	&2.59	&2.03	&3.93	&2.93	 & 48.09\\
  & vicuna-7b-v1.5-16k &2.40	&1.77	&1.40	&3.08	&2.16 & 44.94\\
 & vicuna-13b-v1.5-16k &2.91	&2.12	&1.54	&3.27	&2.46 & 45.39\\
 & Qwen1.5-7B-Chat &3.05	&2.37	&1.88	&3.83	&2.78	& 44.04\\
 & Qwen1.5-14B-Chat &3.29	&2.70	&2.21	&4.04	&3.06	 & 47.42\\
 & Qwen1.5-72B-Chat &3.67	&2.97	&2.98	&\underline{4.21}	&3.46	 & \underline{54.61}\\
  & GPT-3.5-Turbo &3.29	&2.87	&2.17	&3.90	&3.06	 & 51.69\\
 & GPT-4-Turbo   &\textbf{3.81}	&\textbf{3.48}	&\underline{3.36}	&\textbf{4.23}	&\textbf{3.72} & 53.71\\
\midrule
\addlinespace[0.05cm]
\rowcolor[gray]{0.95} \multicolumn{8}{c}{\textit{\dataset (Short subset)}}\\
\midrule
\addlinespace[0.05cm]
\multirow{2}{*}{\textbf{Sum-in-One-Go}}
& GPT-4-Turbo & \textbf{3.98}	&\textbf{3.83}	&\textbf{3.72}	&\textbf{4.28}	&\textbf{3.95}  & \underline{56.79}  \\
& Claude3-Sonnet &\underline{3.81}	&3.32	&3.57	&\underline{4.11}	&\underline{3.70}  & \textbf{61.11}\\
\textbf{Incremental} & GPT-4-Turbo &3.66	&3.47	&\underline{3.62}	&3.72	&3.62  & \textbf{61.11}\\

\textbf{Hierarchical} & GPT-4-Turbo &3.66	&\underline{3.62}	&3.38	&4.09	&3.69  & 51.85\\
\bottomrule
\end{tabular}
\caption{Results of different LLMs performance on character profiling and motivation recognition. 
The abbreviations used in this table stand for the following terms: `Attr' represents `Attributes'; `Rela' stands for `Relationships'; `Even' denotes `Events'; `Pers' indicates `Personality'; `Avg.' refers to the mean values for the scores across the four dimensions. The best scores are \textbf{bolded} and the second best scores are \underline{underlined}.}
\label{tab:main_result}
\vspace{-0.2cm}
\end{table*}

\begin{table}[t]
\centering  
\small
\begin{tabular}{clcc}
\toprule

\textbf{Reasoned by} & \textbf{Ablation Dimension} & \textbf{Acc. $\%$} & \textbf{Std. $\%$}\\

 \midrule
\addlinespace[0.05cm]
 \rowcolor[gray]{0.95} \multicolumn{4}{c}{\textit{Generated Profile (GPT-4-Turbo + incremental updating)}} \\
 \midrule
\addlinespace[0.05cm] 
\multirow{8}{*}{\makecell{GPT-4-Turbo}}
& -                    & \underline{57.75} & 0.32\\
 & Attr          & 57.38 & 0.11\\
& Rela        & 57.30 & 0.37\\
 & Even              & 48.54 & 0.32\\
 & Pers         & 57.08 & 0.31\\
 & Attr\&Rela          & 56.93 & 0.28\\
 & Attr\&Rela\&Even    & 42.62	&0.56\\
 & Attr\&Rela\&Even\&Pers & 40.90 & 0.73 \\
 \midrule
\rowcolor[gray]{0.95} \multicolumn{4}{c}{\textit{Reference Profile in \dataset}} \\
\midrule
\addlinespace[0.05cm]
\multirow{1}{*}{GPT-4-Turbo} & - &  \textbf{63.07} & 0.11\\
\quad human & - & 72.58 & 3.32 \\
\bottomrule
\end{tabular}
\caption{Results of Motivation Recognition ablation study.
The reasoning model by default is GPT-4-Turbo-0409.
\textbf{Ablation Dimension} refers to omitted dimensions in experiments.}
\label{tab:motivation_ablation}
\end{table}
\newcommand{\reditalic}[1]{{\color{red} \textit{#1}}} 
\newcommand{\greenitalic}[1]{{\color{blue} \textit{#1}}} 
\begin{table*}[t]
  \centering
  \small
    \begin{tabularx}{\linewidth}{cXX}
    \toprule

    \textbf{Error type} & \textbf{Generated Profile} & \textbf{Golden Profile}\\

    \midrule

    \multirow{3}{*}{{\makecell{\textbf{Character}\\\textbf{Misidentification}}}} 

    & Benjamin's relationships are complex and multifaceted. He is \reditalic{married to Mildred}, a woman of delicate health and refined tastes $\cdots$ & Rask \greenitalic{marries Helen Brevoort}, a woman from an old-money New York family with a similarly reserved personality$\cdots$ \\
    
    \midrule
    \multirow{3}{*}{{\makecell{\textbf{Relationship}\\\textbf{Misidentification}}}} 

    & Benjamin's life takes a dramatic turn when he saves \reditalic{his grandson, Waldo}, during an unexpected home birth & Benjamin's role as a caregiver extends beyond his family when he helps deliver \greenitalic{Waldo Shenkman, his neighbor's son}, in a dramatic home birth$\cdots$ \\
    
    \midrule

    \multirow{10}{*}{{\makecell{\textbf{Omission of}\\\textbf{Key Information}}}} 
    & Bobby Western's relationships are complex, featuring camaraderie with colleagues like Oiler and Red, a controversial bond with his sister, and deep connections with \reditalic{figures such as Heaven, Asher, Granellen}$\cdots$ & Bobby's most significant relationships are with his sister Alicia, who suffers from schizophrenia and eventually dies by suicide, and \greenitalic{his father, a renowned physicist}$\cdots$\\
    \cmidrule(lr){2-3}
    & Avery continues her work, focusing on \reditalic{helping clients like Marissa and Matthew Bishop navigate their marital issues} $\cdots$ Avery encounters various challenges, including dealing with Skylar's unexpected visit $\cdots$& Matthew is orchestrating these events as part of a revenge plot against Marissa and her affair partner, Skip, whom Avery briefly dated$\cdots$ it's orchestrated by a pharmaceutical company, Acelia, seeking \greenitalic{retribution against Avery for whistleblowing} $\cdots$ \\

    \midrule

    \multirow{9}{*}{{\makecell{\textbf{Event}\\\textbf{Misinterpretation}}}} 
    & In the wake of Mildred's death, Benjamin's life takes a turn towards solitude and reflection. He \reditalic{begins to work on his autobiography} with the help of Ida Partenza, a young secretary$\cdots$ & Returning to New York, Rask realizes his wife's death has little impact on his life. He \greenitalic{continues investing} but never replicates his earlier success, returning to the solitary, dispassionate life$\cdots$\\
    \cmidrule(lr){2-3}
    & Millie's history with Enzo and her relationship with Brock add complexity as she aids Wendy in escaping Douglas's control, \reditalic{accidentally killing Douglas} in the process$\cdots$ & Millie ends up shooting a man \greenitalic{she believes to be Douglas} during a violent altercation, only to \greenitalic{discover later that the man was actually Russell Simonds}$\cdots$ \\

    \midrule

    \multirow{9}{*}{{\makecell{\textbf{Character}\\\textbf{Misinterpretation}}}} 
    & Ava is introspective, self-aware, and \reditalic{morally driven}, with a strong desire for acceptance. She's empathetic but guarded, resourceful in adversity, and adept at navigating complex social situations$\cdots$ & Ava is adept at manipulating situations to her advantage, portraying herself as vulnerable to deceive others while secretly harboring a willingness to \greenitalic{commit fraud to achieve her goals}$\cdots$\\
    \cmidrule(lr){2-3}
    & June Hayward is introspective, ambitious, and somewhat cynical. She navigates her literary career with \reditalic{determination and vulnerability, showing resilience} in the face of criticism and a deep appreciation for her moments of success$\cdots$ & June Hayward is characterized by her intense jealousy, ambition, and insecurity. She is \greenitalic{manipulative, willing to betray} close relationships and ethical boundaries to achieve literary success$\cdots$ \\

    \bottomrule

    \end{tabularx}
  \caption{A case study on common errors generated by models in the character profiling task.}
  \label{tab:case_study}
\end{table*}
\paragraph{Metric for FCE: Consistency Score}
\label{sec:ca_metric}
As demonstrated in a previous study~\cite{goyal2022news}, current reference-based automatic metrics like ROUGE metric~\cite{lin2004rouge} exhibit a significantly low correlation with human judgment for summaries generated by GPT-3.
Therefore, we adopt the evaluation method used in recent research~\cite{liu2023gpteval,gao2023human,li2024translate}, utilizing an LLM as an evaluator to improve alignment with human perception and reduce cost.
\footnote{The result of existing evaluation metrics is provided in Appendix \ref{appendix:trad_metrics}.}
Specifically, we introduce \textbf{Consistency Score}, which is the degree of factual consistency between the reference profiles and the summaries generated by LLMs, evaluated by \llama.
We ask \llama to assign a score on a scale from 1 to 5, reflecting the accuracy of the summaries in capturing the essential factual details. 
A higher score indicates a closer match to the factual content.

To evaluate the quality of the LLM evaluation, we randomly select 50 samples for human evaluation. 
We calculate the Pearson Correlation Coefficient~\cite{cohen2009pearson} between the consistency score result of human annotators and \llama. 
The coefficient value of $0.752$ with the $p\text{-value}=4.3\mathrm{e}{-12}<0.05$ suggests that these two sets of results have a significant correlation.
This validates that the evaluation capabilities of \llama for this task are comparable to those of humans.

\paragraph{Metric for MR: Accuracy}
Multiple-choice questions can be easily evaluated by examining the choice of models. 
We define \texttt{Acc} as the accuracy across the entire question dataset.

\subsection{\dataset Dataset Construction}

\paragraph{Book Dataset}
To reduce the confounding effect of book memorization on the results, we select 126 high-quality novels published in 2022 and 2023.\footnote{Details on the construction process and integrity verification experiments of the \dataset dataset can be found in Appendix \ref{appendix:dataset}.}
In fact, as shown in Appendix \ref{appendix:verification}, we find that there is no significant correlation between the year of publication and the consistency score for works from the past ten years.
For each novel, we concentrate solely on its main character.
We manually remove sections not pertinent to the novel's original content, such as prefaces, acknowledgments, and author introductions.
Additionally, we select 47 books within \dataset containing fewer than 120,000 tokens for the summarizing-in-one-go method.

\paragraph{Golden Character Profile Extraction}
The golden character profiles are gathered from the \method website, known for its high-quality plot summaries and character analyses conducted by literary experts.
With permission from the site, we utilize their book summaries, chapter summaries, and character analyses.
The original character analyses from \method lack a standardized format and predefined profile dimensions.
Therefore, we utilize \gptfour to reorganize the original summaries.

Given the original plot summaries and character analyses, we require the model to reorganize character profiles across four main dimensions while ensuring no critical details are overlooked.
To guarantee the quality of the reorganized profiles, two annotators evaluate whether the reorganized profiles adequately retained the essential information from the original text. 
The assessment reveals that all results exhibit a high level of informational integrity and consistency, confirming the credibility of the reorganized profiles.\footnote{The detail of human examination is shown in Appendix \ref{appendix:reference_human}.}

\paragraph{MR Dataset Construction}
\label{sec:mgchoice}
Using resources from the \method website, we develop motivation recognition questions for key characters in \dataset.
The process involves four main steps:
First, we utilize \gptfour to generate several motivation recognition multiple-choice questions (MCQs) and manually select the top 10 examples.
Second, we identify a primary character from each of the 126 books and formulate questions related to them.
Given the character's name, chapter summaries from the \method, and the 10 examples, \gptfour is instructed to generate a set of motivation recognition multiple-choice questions.
Each question is designed to include a decision made by the character within a specific scenario, offering four options, the correct option, and justifications for why each option is correct or incorrect. 
Through this process, \gptfour generates a total of 641 questions for the 126 characters.
Moreover, we find that some questions can be easily answered using commonsense knowledge or grammatical structure,
Thus, given a question and the correct answer, we ask GPT-4 to provide three likely motivations behind the decision in the question that differ from the correct answer. 
These options, meant to confuse, are similar to the correct answer in sentence structure.
We replace the incorrect options generated in the previous step with these three motivations.

To maintain the quality of MR questions, two annotators are assigned to filter them, with Fleiss's $\kappa=0.91$~\cite{fleiss1981measurement}. 
According to the annotation results, 445 out of the 641 questions meet the established criteria, guaranteeing the quality of the MR questions dataset.\footnote{Further details are shown in Appendix~\ref{appendix:human}.}

\section{Experiment Settings}
\label{sec:exevalu}
\label{setting}
\paragraph{Models for Summarization}
For the incremental and hierarchical methods, we experiment with the following LLMs: Mistral-7B-Instruct-v0.2~\cite{jiang2023mistral}, Mixtral-8x7B-MoE~\cite{jiang2024mixtral}, Qwen1.5-7B-Chat, Qwen1.5-14B-Chat, Qwen1.5-72B-Chat~\cite{bai2023qwen}, vicuna-7b-v1.5-16k, vicuna-13b-v1.5-16k~\cite{zheng2024judging}, GPT-3.5-Turbo-0125 and GPT-4-Turbo-0125. 
We set the chunk size to 3000 tokens for both methods. 
We require that the complete profile generated by the model contain no more than 1200 words.
For the summarizing-in-one-go method, we experiment with the GPT-4-Turbo-0125 and Claude-3-Sonnet~\cite{anthropic2024claude3}.
For all these models, we all adopt the original model and official instruction formats.
The temperatures of all these models are set to 0 in our experiments.

\paragraph{MR Task Setting}
We assess the quality of profiles summarized under different models and methods through the accuracy rate on MR tasks. We uniformly employ GPT-4-Turbo-0409 as the reasoning model for this specific task.
Furthermore, we study human performance in the MR task supported by reference profile in \dataset dataset. We employ two human annotators to answer all the questions and calculate the average accuracy and standard deviation.

\paragraph{Dimension Ablation Study}
To further explore the impact of different dimensions of character information on the MR task, we conduct an analysis through ablation experiments as shown in Table \ref{tab:motivation_ablation}, using character profiles summarized via the incremental method with \gptfour.
Each experiment is repeated three times, and we report the average and standard deviation of the results.

\section{Experiment Results}
\label{sec:analysis}

In the experiments, we wish to answer two research questions:
\begin{inparaenum}[\it RQ1)]
    \item Can LLMs generate character profiles from fictional works precisely?
    \item Can LLMs recognize the character's motivation for a specific decision based on the character profile?
\end{inparaenum}

\subsection{\textit{Can LLMs generate character profiles from fictional works precisely?}}

Experiment result in Table~\ref{tab:main_result} shows that:
\begin{inparaenum}[\it 1)]
    \item LLMs generally exhibit promising performance in generating character profiles from fiction. 
    Among all models, \gptfour consistently outperforms other models across various methods, exhibiting the advanced capability of LLMs to accurately summarize character profiles. 
    \item Despite \gptfour, larger and more complex LLMs, such as Qwen1.5-72B-Chat, tend to achieve higher consistency scores.
    \item There are variations in model performance across different dimensions.
    For example, LLMs typically achieve higher consistency scores in capturing personality but are less effective at summarizing event-related information.
\end{inparaenum}

\paragraph{Summarization Method Comparison}
We compare the outcomes of the incremental and hierarchical methods across the full \dataset. 
For 47 books containing fewer than 120,000 tokens in \dataset, we include the summarizing-in-one-go method.

The results in Table~\ref{tab:main_result} show that the summarizing-in-one-go method achieves the highest consistency scores in all dimensions, surpassing methods that process content in segments. 
We believe this success stems from processing the entire content of a book at once, which maintains the narrative's coherence and minimizes information loss. 
Additionally, since character details are unevenly distributed throughout fiction, summarizing the text in one step allows the model to focus more effectively on the essential elements of the narrative.

The incremental updating method, while slightly lagging in average consistency, performs better in events than hierarchical summarizing.
This performance can be attributed to its iterative updating nature, which allows the model to refine and update its understanding as more information becomes available or as errors are corrected in subsequent passes.
This finding aligns with those reported by \citet{chang2023booookscore}, which indicate that book summaries generated by the incremental method surpass those produced by the hierarchical method in terms of detail.
\paragraph{Error Analysis}
  
We conduct a case study to further investigate why LLMs fail to generate the correct character profile.  
We define five types of errors, \ie, 
1) \textit{Character Misidentification}, which occurs when characters are mistaken for one another, leading to confusion about their actions or roles. 
2) \textit{Relationship Misidentification}, an error where the type of relationship between characters is inaccurately represented.
3) \textit{Omission of Key Information}, a common error where the significant relationships or events are overlooked while less important information is described in excessive detail.
4) \textit{Event Misinterpretation}, events are incorrectly interpreted, or earlier interpretations are not adequately revised in light of subsequent revelations.
5) \textit{Character Misinterpretation}, where the motives or traits of a character are incorrectly summarized, resulting in a cognitive bias in the understanding of a character's overall image.

As shown in Table~\ref{tab:case_study}, a key finding is that the model often becomes confused and generates illusions when faced with complex narrative structures.
For example, in the book ``Trust'', the character Benjamin Rask is a figure in the novel ``Bonds'' which is part of ``Trust''. 
The prototype for Rask is another character, Andrew Bevel, from ``Trust''. 
Due to frequent shifts in narrative perspective, the model confuses Rask with Bevel, mistakenly attributing Bevel’s traits to Rask. 
The errors are shown in the first examples of \textit{Character misidentification} and \textit{Event Misinterpretation}.
Another example occurs in ``The Housemaid's Secret'', where the model fails to understand the plot twist, which results in an incorrect final summary. 
This error is shown in the second case of \textit{Event Misinterpretation}.

\subsection{\textit{Can LLMs recognize the character's motivation for a specific decision?}}

\paragraph{Overall Performance}
As shown in Table~\ref{tab:main_result}, profiles generated by \gptfour through incremental method enable the model to achieve the highest accuracy ($57.75\%$), which is slightly lower than that of the reference profiles ($63.07\%$) shown in Table~\ref{tab:motivation_ablation}, indicating the effectiveness of the generated profiles in enhancing character comprehension.
Additionally, based on the human annotators' results (72.58\%), \gptfour still shows a performance gap compared to humans in this task.

Moreover, a strong positive correlation is observed between the consistency scores and the MR accuracy of the profiles summarized by the model. 
This finding supports the validity of character profiling, suggesting that accurate character profiles help models better understand the motivations behind a character's behavior. 

Among the three summarization methods, profiles from hierarchical merging exhibit relatively low accuracy on the MR task. 
It is also found that despite high scores in other dimensions, the consistency score for the events obtained through the hierarchical method is relatively low. 
This indirectly suggests that the quality of events has a more significant influence on the MR task.

\paragraph{Ablation Study on MR}
As Table~\ref{tab:motivation_ablation} demonstrates, the results of the ablation experiments reveal that each of the four dimensions within the profile contributes to the downstream task. 
Among these, the dimension of the event is the most critical. 
Excluding this dimension alone leads to a notable decrease in accuracy ($-9.21\%$).
The rationale behind this is that events contain substantial plot-related information, which assists the model in grasping the background knowledge pertinent to the characters' decision-making processes. 
Additionally, events integrate elements from the other dimensions, offering a holistic depiction of character personas. However, omitting the other dimensions has a less pronounced impact.
We also observe that reducing the amount of information in the profile correlates with greater variance in experimental outcomes, suggesting that the model becomes less stable as it processes less detailed profiles.

\section{Conclusion}
\label{sec:conclusion}
\vspace{-0.2cm}
We introduce the first task for assessing the character profiling ability of large language models (LLMs), using a dataset of 126 character profiles from novels. 
Our evaluation, which includes the \textit{Factual Consistency Examination} and \task, reveals that LLMs generally perform well.
However, even the most advanced models occasionally generate hallucinations and errors, particularly with complex narratives, highlighting the need for further improvement.

\section*{Limitations}
\label{sec:limitation}
In this paper, we only explore four common dimensions for character profiles, thus leaving other potential dimensions unexplored. 
This limitation suggests that future work could expand the scope to include a wider range of dimensions and investigate their effects on downstream tasks.

Another limitation of our work stems from potential biases in the evaluation process.
Despite selecting highly contemporaneous data to prevent data leakage, it is still possible that some models might have been trained on these specific books.
Besides, the evaluation metrics used in this paper rely on the evaluator LLMs, potentially compromising the accuracy of the results due to errors inherent in these models, which could result in a biased estimation of profile consistency.
Moreover, while we test the three most popular summarization methods, we acknowledge that there is potential for improvement in the design of these methods to maximize the character profiling capabilities of LLMs.

\section*{Ethics Statement}
\label{sec:Ethics}
\paragraph{Use of Human Annotations}
Our institution recruits annotators to implement the annotations of motivation recognition dataset construction. 
We ensure the privacy rights of the annotators are respected during the annotation process.
The annotators receive compensation exceeding the local minimum wage and have consented to the use of motivation recognition data processed by them for research purposes. Appendix~\ref{appendix:human} provides further details on the annotations.

\paragraph{Risks}
The \dataset dataset in our experiment is sourced from publicly available sources. However, we cannot guarantee that they are devoid of socially harmful or toxic language. 
Furthermore, evaluating the data quality of the motivation recognition dataset is based on common sense, which can vary among individuals from diverse backgrounds.
We use ChatGPT~\cite{openai2022chatgpt} to correct grammatical errors in this paper.

\section*{Acknowledgements}
We thank the anonymous reviewers for their valuable suggestions.
This work was supported by the Chinese NSF Major Research Plan (No.92270121).

\bibliography{anthology,custom}

\clearpage
\appendix

\section{Profiling v.s. Other Character Centric Summaries}
\label{appendix:comparison}
\subsection{Dimension}
Some previous work also summarizes certain information about characters from books or scripts.
However, these studies on character understanding tend to focus on one specific aspect of the character, such as role~\cite{stammbach2022heroes}, relationship~\cite{Zhao2024LargeLM}, personality~\cite{yu2023personality},
mental states~\cite{yu2022few}.
Furthermore, although these studies offer valuable insights into character understanding, their specific focus may not capture the multifaceted nature of character.
Although recent RPA works have managed to summarize character information in a multi-dimensional approach, such as attributes, appearance, relationship, storyline~\cite{zhao2023narrativeplay,li2023chatharuhi,wang2023rolellm}, there is a lack of systematic assessment of the quality of these summaries.

\subsection{Evaluation}
Our evaluation framework of character profiling covers a wide range of information related to characters, facilitating a multidimensional understanding of characters.
Although we explicitly request models to consider four dimensions, many dimensions of information are included in our framework or can be easily derived from the summarized profile. 
Specifically, in contrast to key mental states~\cite{yu2022few}, our framework inherently encompasses critical mental information.
For example, a character's objectives, part of their attributes profile defined in Section~\ref{sec:dims}, reflect their desires and intentions. Key emotions and beliefs are often revealed through their reactions and behaviors in the profile of the events.
Our work also includes factual details, like relationships with other characters and interactions with the external world.

\subsection{Application}
We believe that the extensive character summary can provide necessary and valuable information for various downstream applications, \eg chatbots of fictional characters~\cite{chen2023large}, interactive narratives~\cite{zhao2023narrativeplay}, and study guides for human readers.
However, since we prompt the model to limit the total word count of the entire profile, some dimensions may be more concise and not as detailed as summaries that focus solely on that dimension.

\section{\dataset Dataset}
\label{appendix:dataset}
\subsection{Dataset Construction}
We select 126 books to construct our dataset.
For each book, we collect the book's epub format and transform it into TXT format, and then process the texts into chunks of content with the required chunk size.
All 126 books are fictional novels with an average token count of 134412. 
Among these books, 47 books are less than 120k tokens in length, and the average token count of these books is 101885.

To minimize the potential for data leakage, we exclusively restrict our book selection to those published within the years 2022 and 2023.
Additionally, we ensure that the selected books are either not sequels or, if they are sequels, can be regarded as independent works.
Our selection focuses on works of fiction, specifically excluding biographical novels and other works based on real historical figures.

For the evaluation of our work, we obtain permission from the developer of the \method website to use the summaries and character analyses of these books written by experts.
All book summaries and character analyses are used for academic research purposes. 
We release only the reorganized profiles, not the original summaries to protect the developers' copyrights.

\subsection{Integrity Verification of \dataset dataset}
\label{appendix:verification}

To confirm that our datasets and task genuinely evaluate understanding capability rather than simply testing the recovery of LLM training data, we design experiments to demonstrate that data leakage is not a significant concern with our dataset.
In the following four datasets, all profiles are generated by \gptfour through the incremental updating method.

\paragraph{For Publication Years}
We count the number of books in \dataset dataset published in the years 2022 and 2023, and their average consistency scores are shown in Table \ref{tab:year_ablation}. 
The average consistency scores of the books from 2022 and 2023 are very close ($3.60$ vs $3.62$).

\paragraph{For Different Sales Volumes}
We collect the collections tag in \method for books in \dataset dataset. The results of books in the ``New York Times Best Sellers'' collection and those that are not are shown in Table \ref{tab:bestseller_ablation}. 
These two consistency scores are also very close ($3.58$ v.s. $3.62$).
The two experiments above demonstrate that within \dataset dataset, publication year and sales volume do not significantly affect task performance. 
Furthermore, we collect a small set of highly canonized texts for comparison with our dataset.

\paragraph{For Classic Works in the 20th Century} 
We gather the top 10 books (excluding series) from the ``Best Books of the 20th Century'' list on the \textit{goodreads}\footnote{\url{https://www.goodreads.com}} website. 
This collection includes well-known classics like ``The Little Prince'', ``1984'', and ``One Hundred Years of Solitude''.
The results of this set are shown in Table \ref{tab:classic}, where the consistency score is much higher than that of \dataset dataset. 
This finding suggests that our selection of data effectively reduces the impact of data leakage compared with choosing classic works.
We believe that high performance is due to the accumulation of time. The training set contains a large number of related corpora, such as Wikipedia entries, literary analyses, fan creations, and other related content, which deepen the model's understanding of these books and characters.

\begin{figure}[t]
    \centering
    \includegraphics[width=\linewidth]{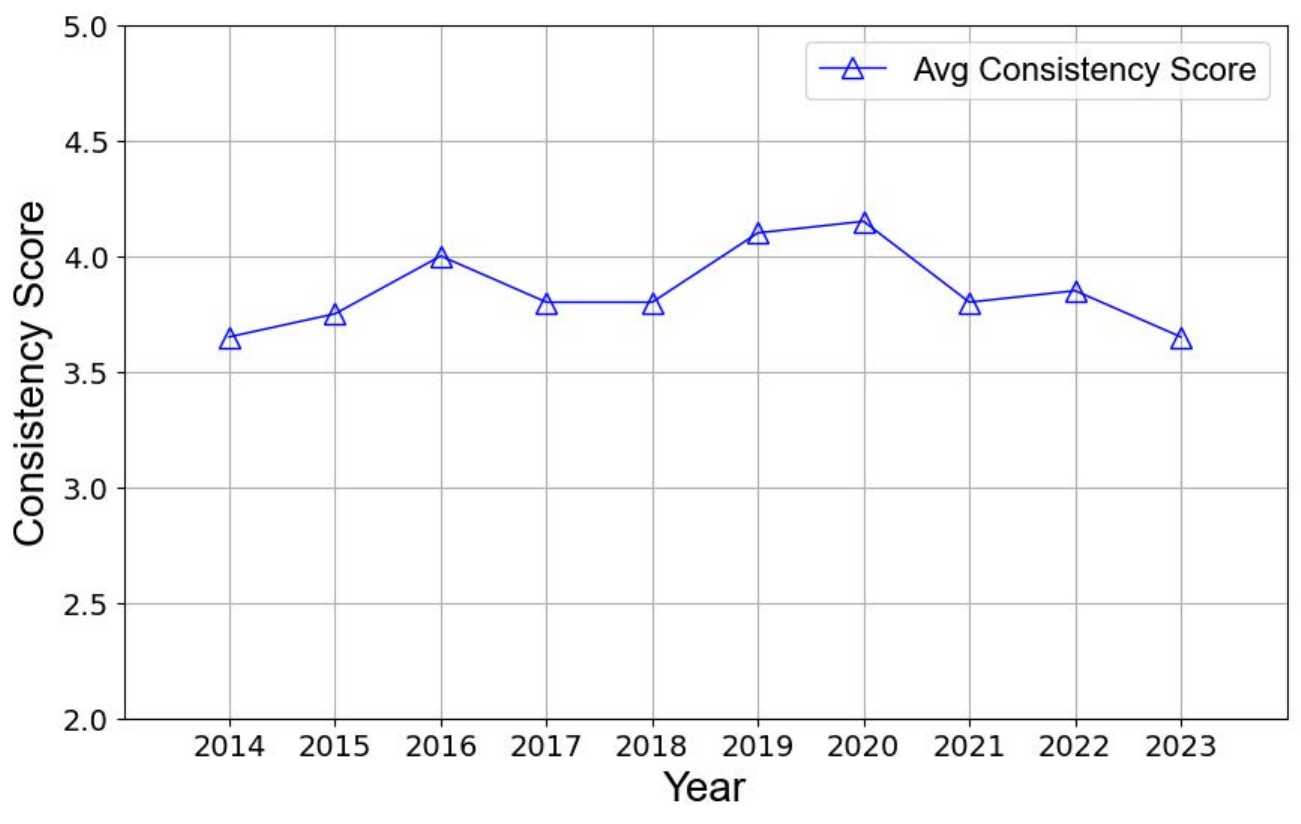}
    \caption{Average Consistency Score of books in ``Best Books in \{\#the year\}'' list in \textit{goodreads} in different years.}
    \label{fig:year-plot}
\end{figure}

\paragraph{For Books Over Last Ten Years}
In order to test the impact of publication year on task performance in more recent books, we collect books from the ``Best Books in \{\#the year\}'' list on the \textit{goodreads} website.
We gather five books published in each of the last ten years, making a total of fifty books.
The results of the average consistency score over different years are shown in Figure \ref{fig:year-plot}, and the detailed scores on different dimensions are shown in Table \ref{tab:year_plot}.
We conduct a Spearman Rank Correlation Coefficient Test~\cite{spearman1961proof} on this set.
The coefficient of $-0.037$ with $p$-value of $0.799$ (>$0.05$) indicates no significant correlation between the year of publication and the average consistency score over these 50 samples.
This result suggests that even though the texts of books from a few years ago may well have been trained by the model, there are not enough related corpora to allow the model to perform well on this task solely based on memory.

Based on the above analysis, we reasonably speculate that, at least for the next few years, our dataset will remain effective for updated LLMs.
Moreover, this work does not only focus on the dataset itself but, importantly, on a feasible framework designed to continually update and expand this dataset. 
Furthermore, we will keep updating the dataset and evaluating the performance of new LLMs in our future works.

\newcolumntype{a}{>{\columncolor{BlueGreen!10}}c}
\newcolumntype{b}{>{\columncolor{brown!10}}r}
\newcolumntype{d}{>{\columncolor{blue!10}}r}
\newcolumntype{q}{>{\columncolor{Green!10}}r}
\begin{table}[t]
\centering  
\footnotesize
\begin{tabular}{ccddddb}
\toprule
\multicolumn{1}{c}{\multirow{2}[1]{*}{\makecell{\textbf{Year}}}} &  
\multicolumn{1}{c}{\multirow{2}[1]{*}{\textbf{Count}}} &
\multicolumn{5}{c}{\textbf{Consistency Score}} \\

\cmidrule(lr){3-7}
& & \multicolumn{1}{c}\textbf{Attr} & \multicolumn{1}{c}\textbf{Rela} & \multicolumn{1}{c}\textbf{Even} & \multicolumn{1}{c}\textbf{Pers} & \multicolumn{1}{c}\textbf{Avg.}\\
\midrule
\addlinespace[0.05cm]
2022 & 93 &3.70	&3.23	&3.71	&3.75	&3.60 \\
2023 & 33&3.79	&3.27	&3.21	&4.21	&3.62\\
\bottomrule
\end{tabular}
\caption{Results of character profiling on books published in different years.}
\label{tab:year_ablation}
\vspace{-0.2cm}
\end{table}

\newcolumntype{a}{>{\columncolor{BlueGreen!10}}c}
\newcolumntype{b}{>{\columncolor{brown!10}}r}
\newcolumntype{d}{>{\columncolor{blue!10}}r}
\newcolumntype{q}{>{\columncolor{Green!10}}r}
\begin{table}[t]
\centering  
\footnotesize
\begin{tabular}{ccddddb}
\toprule
\multicolumn{1}{c}{\multirow{2}[1]{*}{\makecell{\textbf{Bestseller}}}} &  
\multicolumn{1}{c}{\multirow{2}[1]{*}{\textbf{Count}}} &
\multicolumn{5}{c}{\textbf{Consistency Score}} \\

\cmidrule(lr){3-7}
& & \multicolumn{1}{c}\textbf{Attr} & \multicolumn{1}{c}\textbf{Rela} & \multicolumn{1}{c}\textbf{Even} & \multicolumn{1}{c}\textbf{Pers} & \multicolumn{1}{c}\textbf{Avg.}\\
\midrule
\addlinespace[0.05cm]
Yes & 43 &3.67	&3.28	&3.44	&3.91	&3.58 \\
No & 83 &3.75	&3.22	&3.65	&3.86	&3.62\\
\bottomrule
\end{tabular}
\caption{Results of character profiling on books in ``New York Times Best Sellers'' collection in \method and those are not.}
\label{tab:bestseller_ablation}
\vspace{-0.2cm}
\end{table}

\newcolumntype{a}{>{\columncolor{BlueGreen!10}}c}
\newcolumntype{b}{>{\columncolor{brown!10}}r}
\newcolumntype{d}{>{\columncolor{blue!10}}r}
\newcolumntype{q}{>{\columncolor{Green!10}}r}
\begin{table}[t]
\centering  
\footnotesize
\begin{tabular}{cddddb}
\toprule
 
\multicolumn{1}{c}{\multirow{2}[1]{*}{\textbf{Count}}} &
\multicolumn{5}{c}{\textbf{Consistency Score}}\\

& \multicolumn{1}{c}\textbf{Attr} & \multicolumn{1}{c}\textbf{Rela} & \multicolumn{1}{c}\textbf{Even} & \multicolumn{1}{c}\textbf{Pers} & \multicolumn{1}{c}\textbf{Avg.}\\
\midrule

\addlinespace[0.05cm]
\rowcolor[gray]{0.95} \multicolumn{6}{c}{\textit{Best Books of the 20th century}}\\
\midrule
10 &4.7	&4.1	&4.2	&4.8	&4.45 \\
\bottomrule
\end{tabular}
\caption{Results of character profiling on 10 books in ``Best Books of the 20th Century'' list in \textit{goodreads} .}
\label{tab:classic}
\vspace{-0.2cm}
\end{table}

\newcolumntype{a}{>{\columncolor{BlueGreen!10}}c}
\newcolumntype{b}{>{\columncolor{brown!10}}r}
\newcolumntype{d}{>{\columncolor{blue!10}}r}
\newcolumntype{q}{>{\columncolor{Green!10}}r}
\begin{table}[t]
\centering  
\footnotesize
\begin{tabular}{ccddddb}
\toprule
\multicolumn{1}{c}{\multirow{2}[1]{*}{\makecell{\textbf{Year}}}} &  
\multicolumn{1}{c}{\multirow{2}[1]{*}{\textbf{Count}}} &
\multicolumn{5}{c}{\textbf{Consistency Score}} \\

\cmidrule(lr){3-7}
& & \multicolumn{1}{c}\textbf{Attr} & \multicolumn{1}{c}\textbf{Rela} & \multicolumn{1}{c}\textbf{Even} & \multicolumn{1}{c}\textbf{Pers} & \multicolumn{1}{c}\textbf{Avg.}\\
\midrule
\addlinespace[0.05cm]
2023 & 5 &3.4	&3.8	&3.2	&4.2	&3.65 \\
2022 & 5 &4.2	&3.2	&3.6	&4.4	&3.85\\
2021 & 5 &4.0	&3.4	&3.6	&4.2	&3.80\\
2020 & 5 &4.4	&4.0	&3.8	&4.4	&4.15\\
2019 & 5 &4.0	&3.8	&4.0	&4.6	&4.10\\
2018 & 5 &3.8	&3.6	&4.0	&3.8	&3.80\\
2017 & 5 &3.4	&4.4	&3.2	&4.2	&3.80\\
2016 & 5 &4.4	&3.6	&4.0	&4.0	&4.00\\
2015 & 5 &4.0	&3.6	&3.4	&4.0	&3.75\\
2014 & 5 &4.0	&2.8	&3.8	&4.0	&3.65\\
\bottomrule
\end{tabular}
\caption{Results of character profiling on books in ``Best Books in \{\#the year\}'' list in \textit{goodreads}.}
\label{tab:year_plot}
\vspace{-0.2cm}
\end{table}

\section{Detailed Information of Summarization Method}

Given a book $B$ with length $L$, for chunk-based method, we split $B$ into independent chunk $c_1, c_2, \cdots, c_{\lceil L/C \rceil}$ with chunk size $C=3000$.
We fix the context window $W=8096$ and the maximum summary length $M=1200$.

\subsection{Incremental Updating}
The progress of incremental updating is listed as follows:
\begin{itemize}[noitemsep, leftmargin=*]
    \item Step 1: Given the first chunk $c_1$, the model outputs the initial summary $s_1$.
    \item Step 2: Given the chunk content $c_2$, and the summary $s_1$, the model outputs summary $s_2$ which contains content of the first two chunk.
    \item The summary is iteratively updated within the next chunk through step 2 until the final summary $s_{\lceil L/C \rceil}$ is obtained.
    \item If the summary exceeds $M$ in these steps, the model is required to compress the summary into the required length.
\end{itemize}
\subsection{Hierarchical Merging}
The progress of hierarchical merging is listed as follows:
\begin{itemize}[noitemsep, leftmargin=*]
    \item Step 1: Given the chunks $c_1, c_2, \cdots, c_{\lceil L/C \rceil}$, the model outputs the level 1 summaries for each chunk.
    \item Step 2: Merge as many consecutive level 1 summaries as possible with the limit that the total length of the summaries and the prompt is less than $W$. 
    Given these summaries, the model outputs the first level 2 summary, which serves as the context for the next merging.
    \item Step 3: Merge as many remaining level 1 summaries as possible with the limit that the total length of these summaries and the prompt and the context is less than $W$. Given this input, the model outputs the next level 2 summary, which also serves as the context for the next merging. This process is iteratively conducted within the remaining summaries.
    \item Merge the level 2 summaries by repeating steps 2 and 3 until a final summary is obtained.
\end{itemize}

\subsection{Summarizing in One Go}
We first ensure the total length of the selected book and the summarizing prompt is less than the context window limit of \gptfour and Claude-3-Sonnet.
Given the whole content of the book, the model outputs the final summary at once.

\section{Manual Annotation}
We invite four native English-speaking college students as human annotators for manual evaluation in our work.
These annotators receive compensation exceeding the local minimum wage.
Two annotators also have consented to the use of motivation recognition data filtered by them for research purposes.
\label{appendix:human}
\subsection{Reference Profile Examination}
\label{appendix:reference_human}
To examine the correctness of the character profile reorganized by GPT-4 from the original book summary and character analysis, we employ two annotators to check the consistency between the reorganized profile and the original content.
The annotators are given the origin plot summary, character analysis, and reorganized character profile.
Then they are required to determine whether the reorganized profile is consistent with the original information.
The two annotators' result shows that the profiles of all samples in \dataset dataset do not contain plot inconsistencies and misjudgments of the character's traits.
This result indicates that the profile's quality meets the standard of a golden profile.

\subsection{Manual Evaluation of FCE}
In order to examine the quality of \llama evaluator result, we sample 50 pieces in our dataset and invite two annotators to evaluate the generated profile in consistency score.
We provide the annotators and \llama with the same scoring prompt. 
For the metric consistency score, the Pearson Correlation Coefficient between the average human result and \llama scoring is 0.752 with $p\text{-value}=4.3\mathrm{e}{-12}$.
The $p$-value $<0.05$ demonstrates that these two sets of results have a significant correlation.
The coefficient result indicates that the \llama evaluation ability is comparable with human annotators on the assessing character profile.

\subsection{Motivation Recognition MCQs filtering}
To ensure the quality of the MR question dataset, we employ two annotators for conducting manual filtering.
The annotators are provided with reference character profiles, generated questions, and the following criteria:

\begin{itemize}[noitemsep, leftmargin=*]
    \item 
    \textbf{The decision must be made by the selected character.}
    Each question must feature a decision and the scenario, with the focus character as the decision-maker.
    \item \textbf{Questions should ask directly or indirectly about the character's motivation for making the decision. }
    Each question must directly or indirectly inquire about the character's motivation for making their decision, avoiding irrelevant information.
    \item \textbf{The decision must be meaningful within the story context.}
    The decision in the question must contribute meaningfully to the storyline.
    It should reflect a conscious choice by the character that holds importance in the narrative, rather than representing a mundane or routine decision.
    \item \textbf{Leaking questions is prohibited.} Scenarios and questions must not include the motivation behind the characters' decisions.
\end{itemize}

We require the annotators to determine if the question meets the criteria. 
By filtering the dataset, we finally get 445 high-quality motivation recognition multiple-choice questions with Fleiss's $\kappa=0.91$.
We also adjust the arrangement of the options to ensure a fair distribution of correct answers.

\subsection{Human Performance in Motivation Recognition}
We employ two annotators to study the human performance in the MR task. The annotators are provided with the complete information in the MR dataset in \dataset except for the correct answer and the reason. They are instructed to choose an option for each question.

\section{Traditional Metrics on Generated Profiles}
\label{appendix:trad_metrics}
In our evaluation protocol, traditional metrics for text summarization like ROUGE~\cite{lin2004rouge}, BLEU~\cite{papineni2002bleu}, and BERTScore~\cite{zhang2019bertscore} are not used because they have been shown to be unreliable for measuring summary quality of GPT-3 generated summaries compared to human evaluations~\cite{goyal2022news}.
However, to provide a comprehensive perspective, we present the results of these three traditional metrics in Table \ref{tab:trad_metric} and Table \ref{tab:trad_bert} for reference.
To ensure the fairness and accuracy of evaluations using lexical metrics, preventing the language styles of different models from influencing the results, we use the profile extracted from original \method data by \llama rather than \gptfour as reference profile in this section.

The results of traditional metrics are generally consistent with the conclusions of the consistency score, showing similar patterns: 
\begin{inparaenum}[\it 1)]
\item Larger and more complex LLMs perform better; 
\item The performance on the event dimension is often the lowest; 
\item For different summarization methods, the average results show that summarizing in one go performs better than hierarchical summarizing, which in turn performs better than incremental updating.
\end{inparaenum}
It is also shown that the relative performance of models varies widely across different metrics.
We believe this happens because each metric focuses on different aspects.
ROUGE-L and BLEU mostly measure lexical matching, while BERTScore emphasizes semantic similarity, and the consistency score checks factual accuracy.
Since profiles generated by different models can be semantically or factually consistent but differ in wording and syntax, their performance may vary depending on the metric used.

\begin{table}[t]
  \centering
  \small
    \begin{tabularx}{\linewidth}{X}
    \toprule
    \rowcolor[gray]{0.95}\multicolumn{1}{c}{\textbf{Init Feedback (Incremental)}} \\
    If there is no information about character \{\} in the beginning\\ part of a story, just output `None' in each section. Do not\\ apologize. Just output in the required format.\\
    \midrule
    \rowcolor[gray]{0.95}\multicolumn{1}{c}{\textbf{Init Feedback (Hierarchical)}} \\
    If there is no information about character \{\} in this part of the story, just output `None' in each section. Do not apologize. Just output in the required format.\\
    \midrule
    \rowcolor[gray]{0.95}\multicolumn{1}{c}{\textbf{Update Feedback (Incremental)}} \\
    If there is no information about character \{\} in this excerpt,\\ just output the origin summary of the character \{\} of the\\ story up until this point. Do not apologize. Just output in the\\ required format.\\
    \bottomrule
    \end{tabularx}
  \caption{The additional prompt for the \gptfour model.}
  \label{tab:feedback_gpt4}
\end{table}
\newcolumntype{a}{>{\columncolor{BlueGreen!10}}c}
\newcolumntype{b}{>{\columncolor{brown!10}}r}
\newcolumntype{d}{>{\columncolor{blue!10}}r}
\newcolumntype{q}{>{\columncolor{Green!10}}r}
\begin{table*}[t]
\centering  
\footnotesize
\begin{tabular}{llrrrrrrrrrr}
\toprule
\multicolumn{1}{c}{\multirow{2}[1]{*}{\makecell{\textbf{Summarization}\\\textbf{Method}}}} &  
\multicolumn{1}{c}{\multirow{2}[1]{*}{\textbf{Summarization Model}}} &
\multicolumn{5}{c}{\textbf{ROUGE-L} \%} & 
\multicolumn{5}{c}{\textbf{BLEU} \%}\\

\cmidrule(lr){3-7}
\cmidrule(lr){8-12}
& & \multicolumn{1}{c}\textbf{Attr} & \multicolumn{1}{c}\textbf{Rela} & \multicolumn{1}{c}\textbf{Even} & \multicolumn{1}{c}\textbf{Pers} & \multicolumn{1}{c}\textbf{Avg.} & \multicolumn{1}{c}\textbf{Attr} & \multicolumn{1}{c}\textbf{Rela} & \multicolumn{1}{c}\textbf{Even} & \multicolumn{1}{c}\textbf{Pers} & \multicolumn{1}{c}\textbf{Avg.}\\
\midrule
\rowcolor[gray]{0.95} \multicolumn{12}{c}{\textit{\dataset (Full dataset)}} \\
\midrule
\addlinespace[0.05cm]
\multirow{9}{*}{\makecell[c]{\textbf{Incremental}\\\textbf{Updating}}} & Mistral-7B-Instruct-v0.2 & 23.89 & 27.04 & 22.16 & 18.70 & 22.94 & 5.41 & 5.88 & 3.32 & 2.76 & 4.34 \\
 & Mixtral-8x7B-MoE & 25.01 & 27.94 & 20.55 & 19.26 & 23.19 & 5.57 & 6.02 & 3.10 & 3.28 & 4.49\\
 & vicuna-7b-v1.5-16k & 30.53 & 30.46 & 21.69 & 24.04 & 26.68 & 8.48 & 7.24 & 2.65 & 4.32 & 5.67\\
 & vicuna-13b-v1.5-16k & 30.60 & 29.58 & 22.94 & 21.75 & 26.22 & 8.56 & 6.41 & 3.60 & 3.28 & 5.46\\
 & Qwen1.5-7B-Chat & 22.04 & 23.04 & 19.10 & 20.83 & 21.25 & 3.66 & 3.79 & 2.11 & 3.58 & 3.28\\
 & Qwen1.5-14B-Chat & 23.05 & 22.03 & 18.13 & 21.01 & 21.05 & 4.53 & 4.12 & 1.48 & 3.18 & 3.33\\
 & Qwen1.5-72B-Chat & 26.94 & 29.08 & 21.68 & 23.32 & 25.26 & 6.29 & 7.50 & 2.76 & 3.90 & 5.11\\
 & GPT-3.5-Turbo & 29.35 & 27.71 & 20.44 & 21.09 & 24.65 & 7.46 & 6.55 & 2.97 & 3.29 & 5.07\\
 & GPT-4-Turbo & 28.61 & 27.56 & 20.62 & 21.98 & 24.69 & 8.00 & 6.67 & 3.10 & 3.17 & 5.23 \\
\cdashlinelr{1-12}
\multirow{9}{*}{\makecell[c]{\textbf{Hierarchical}\\\textbf{Merging}}} & Mistral-7B-Instruct-v0.2 & 27.94 & 29.23 & 22.64 & 23.12 & 25.73 & 9.07 & 8.57 & 3.81 & 5.51 & 6.74\\
 & Mixtral-8x7B-MoE &26.91 & 30.79 & 24.41 & 21.79 & 25.97 & 6.55 & 8.14 & 4.28 & 3.74 & 5.68 \\
  & vicuna-7b-v1.5-16k & 27.87 & 29.31 & 21.47 & 23.47 & 25.53 & 6.20 & 5.85 & 2.10 & 3.98 & 4.53\\
 & vicuna-13b-v1.5-16k & 28.56 & 29.34 & 22.12 & 22.08 & 25.53 & 7.38 & 6.02 & 2.66 & 3.23 & 4.82\\
 & Qwen1.5-7B-Chat & 25.19 & 26.44 & 19.29 & 22.52 & 23.36 & 5.00 & 5.41 & 2.15 & 4.05 & 4.15\\
 & Qwen1.5-14B-Chat & 24.15 & 24.12 & 19.07 & 21.08 & 22.11 & 4.72 & 5.34 & 1.54 & 2.89 & 3.62\\
 & Qwen1.5-72B-Chat & 29.46 & 31.26 & 25.17 & 24.60 & 27.62 & 8.61 & 8.84 & 4.66 & 5.14 & 6.81\\
 & GPT-3.5-Turbo & 32.05 & 29.98 & 21.58 & 24.46 & 27.02 & 9.33 & 7.56 & 3.06 & 4.17 & 6.03\\
 & GPT-4-Turbo &28.88 &26.01 & 21.85 & 23.03 & 24.94 & 8.42 & 7.06 & 3.59 & 3.45 & 5.63 \\
\midrule
\addlinespace[0.05cm]
\rowcolor[gray]{0.95} \multicolumn{12}{c}{\textit{\dataset (Short subset)}}\\
\midrule
\addlinespace[0.05cm]
\multirow{2}{*}{\textbf{Sum-in-One-Go}}
& GPT-4-Turbo & 33.04 & 33.37 & 25.30 & 25.09 & 29.20 & 11.41 & 10.37 & 4.34 & 4.33 & 7.61\\
& Claude3-Sonnet & 37.21 & 35.12 & 28.22 & 26.01 & 31.64 & 13.33 & 11.57 & 5.80 & 4.40 & 8.78\\
\textbf{Incremental} & GPT-4-Turbo & 29.89 & 28.16 & 21.25 & 22.07 & 25.34 & 8.96 & 7.30 & 3.12 & 3.15 & 5.63\\

\textbf{Hierarchical} & GPT-4-Turbo & 30.01 & 26.76 & 21.75 & 23.26 & 25.44 & 9.17 & 7.67 & 3.63 & 3.59 & 6.02\\
\bottomrule
\end{tabular}
\caption{Metric ROUGE-L and BLEU of different LLMs performance on character profiling.}
\label{tab:trad_metric}
\vspace{-0.2cm}
\end{table*}

\newcolumntype{a}{>{\columncolor{BlueGreen!10}}c}
\newcolumntype{b}{>{\columncolor{brown!10}}r}
\newcolumntype{d}{>{\columncolor{blue!10}}r}
\newcolumntype{q}{>{\columncolor{Green!10}}r}
\begin{table*}[t]
\centering  
\footnotesize
\begin{tabular}{llccccc}
\toprule
\multicolumn{1}{c}{\multirow{2}[1]{*}{\makecell{\textbf{Summarization}\\\textbf{Method}}}} &  
\multicolumn{1}{c}{\multirow{2}[1]{*}{\textbf{Summarization Model}}} &
\multicolumn{5}{c}{\textbf{BertScore \%}} \\

\cmidrule(lr){3-7}
& & \multicolumn{1}{c}\textbf{Attr} & \multicolumn{1}{c}\textbf{Rela} & \multicolumn{1}{c}\textbf{Even} & \multicolumn{1}{c}\textbf{Pers} & \multicolumn{1}{c}\textbf{Avg.}\\
\midrule
\rowcolor[gray]{0.95} \multicolumn{7}{c}{\textit{\dataset (Full dataset)}} \\
\midrule
\addlinespace[0.05cm]
\multirow{9}{*}{\makecell[c]{\textbf{Incremental}\\\textbf{Updating}}} & Mistral-7B-Instruct-v0.2 & 87.25 & 87.95 & 84.88 & 85.97 & 86.51 \\
 & Mixtral-8x7B-MoE & 86.98 & 85.37 & 75.89 & 77.21 & 81.36\\
 & vicuna-7b-v1.5-16k & 87.61 & 87.81 & 84.89 & 85.56 & 86.47\\
 & vicuna-13b-v1.5-16k & 87.62 & 87.67 & 84.91 & 82.92 & 85.78\\
 & Qwen1.5-7B-Chat & 86.80 & 87.26 & 84.73 & 86.98 & 86.44\\
 & Qwen1.5-14B-Chat & 87.28 & 87.51 & 84.88 & 87.43 & 86.78\\
 & Qwen1.5-72B-Chat & 88.05 & 88.78 & 85.48 & 87.80 & 87.53\\
 & GPT-3.5-Turbo & 87.86 & 88.05 & 84.90 & 87.13 & 86.98\\
 & GPT-4-Turbo & 87.91 & 88.27 & 85.32 & 87.33 & 87.21\\
\cdashlinelr{1-7}
\multirow{9}{*}{\makecell[c]{\textbf{Hierarchical}\\\textbf{Merging}}} & Mistral-7B-Instruct-v0.2 &87.95 & 88.46 & 83.95 & 87.48 & 86.96\\
 & Mixtral-8x7B-MoE & 87.58 & 87.63 & 84.74 & 85.96 & 86.47\\
  & vicuna-7b-v1.5-16k & 86.14 & 86.59 & 84.51 & 85.42 & 85.66\\
 & vicuna-13b-v1.5-16k & 87.20 & 87.50 & 85.10 & 86.08 & 86.47\\
 & Qwen1.5-7B-Chat & 87.22 & 87.82 & 84.94 & 87.22 & 86.80\\
 & Qwen1.5-14B-Chat & 87.83 & 88.05 & 85.42 & 87.36 & 87.16\\
 & Qwen1.5-72B-Chat & 88.42 & 89.10 & 86.37 & 88.01 & 87.97\\
  & GPT-3.5-Turbo & 88.74 & 88.74 & 85.39 & 87.86 & 87.68\\
 & GPT-4-Turbo   & 88.38 & 88.63 & 85.58 & 87.73 & 87.58\\
\midrule
\addlinespace[0.05cm]
\rowcolor[gray]{0.95} \multicolumn{7}{c}{\textit{\dataset (Short subset)}}\\
\midrule
\addlinespace[0.05cm]
\multirow{2}{*}{\textbf{Sum-in-One-Go}}
& GPT-4-Turbo & 89.18 & 89.81 & 86.83 & 88.45 & 88.56\\
& Claude3-Sonnet & 89.38 & 89.70 & 87.08 & 88.19 & 88.59\\
\textbf{Incremental} & GPT-4-Turbo & 88.26 & 88.48 & 85.43 & 87.26 & 87.36\\

\textbf{Hierarchical} & GPT-4-Turbo & 88.61 & 88.86 & 85.68 & 87.79 & 87.73\\
\bottomrule
\end{tabular}
\caption{Metric BERTScore of different LLMs performance on character profiling.}
\label{tab:trad_bert}
\vspace{-0.2cm}
\end{table*}

\section{Prompts}
For summarization, we mainly adopt the prompt structure from \citet{chang2023booookscore}.
\subsection{Summarizing in One Go}
In our experiment, we have found that the long-context capabilities of Claude-3-Sonnet are limited. 
Consequently, the model occasionally forgets the instructions and generates a simplistic summary instead of organizing the output into four distinct sections when the task prompt precedes the novel's content. 
Therefore, we choose to put the task prompt after the content of the novel.
The prompts for summarizing-in-one-go method can be found in Table \ref{tab:template_for_extract_at_once}.  

\subsection{Incremental Updating}
The prompts for incremental updating can be found in Table \ref{tab:prompt_for_incremental}.

We have found that the \gptfour model will provide an apology if there is no information available about the designated character in the current excerpt, instead of outputting in the required format.
So we add an additional prompt for the \gptfour model and regenerate, if the response starts with apology. 
The feedback prompt can be found in Table \ref{tab:feedback_gpt4}.

\subsection{Hierarchical Summarizing}
Likewise, we add a feedback prompt for the \gptfour model if the response starts with an apology. The prompts for hierarchical summarizing can be found in Table \ref{tab:prompt_for_hierarchical}.

\subsection{Factual Consistency Examination}
For evaluation, we mainly adopt the prompt structure from \citet{liu2023gpteval}.
The prompt template is shown in Table \ref{tab:prompt_for_fce}.

\subsection{Motivation Recognition}
The prompt template of MR task is shown in Table \ref{tab:prompt_for_mr}.

\begin{table*}[b]
  \centering
  \small
    \begin{tabular}{l}
    \toprule
    \makecell[l]{
    \color{gray}{/* \textit{Data} */}\\
Below is the content of the novel:\\
\\
- - -\\
\{\}\\
- - -\\
    \color{gray}{/* \textit{Task prompt} */} \\
You are a character persona extraction assistant. Your task is to write a summary for the character \{\} in this novel. You\\ must briefly introduce characters, places, and other major elements if they are being mentioned for the first time in the\\ summary. The story may feature non-linear narratives, flashbacks, switches between alternate worlds or viewpoints, etc.\\ Therefore, you should organize the summary so it presents a consistent and chronological narrative. The summary must\\ be within \{\} words and could include multiple paragraphs.\\
\color{gray}{/* \textit{Output Format} */} \\
Output your summary in four specific sections, using the following titles as paragraph headers:\\
\\
Attributes: // Briefly identify the character's gender, skill, talents, objectives, and background within \{\} words.\\
Relationships: // Briefly describe the character's relationships with other characters within \{\} words.\\
Events: // Organize the main events the character experiences or is involved in chronological order within \{\} words.\\
Personality: // Briefly identify the character's personality within \{\} words.\\
\\
Ensure that each section explicitly starts with the specified title, followed by the content and that there is a clear separation\\ (a newline) between each section.\\
\\
Summary:\\ 
\color[rgb]{0,0.39,0}\textit{Attributes:}\\
\color[rgb]{0,0.39,0}\textit{Margot Davies is a determined and skilled female reporter with... }\\
\\
\color[rgb]{0,0.39,0}\textit{Relationships:}\\
\color[rgb]{0,0.39,0}\textit{Margot has a close and loving relationship with her uncle...}\\ 
\\
\color[rgb]{0,0.39,0}\textit{Events:}\\
\color[rgb]{0,0.39,0}\textit{Margot returns to her hometown of Wakarusa to care for her ailing uncle...}\\ 
\\
\color[rgb]{0,0.39,0}\textit{Personality:}\\
\color[rgb]{0,0.39,0}\textit{Margot is tenacious, intelligent, and compassionate....}}\\ 
    \bottomrule
    \end{tabular}
  \caption{Prompt templates for summarizing-in-one-go method. Generated texts by a LLM are {\color[rgb]{0,0.39,0}\textit{highlighted}}.
  }
  \label{tab:template_for_extract_at_once}
\end{table*}
\begin{table*}[t]
  \centering
  \small
    \begin{tabular}{l}
    \toprule
    \rowcolor[gray]{0.95}\multicolumn{1}{c}{\textbf{I: Init}} \\
 \makecell[l]{
    \color{gray}{/* \textit{Data} */}\\
Below is the beginning part of a story:\\
\\
- - -\\
\{\}\\
- - -\\
    \color{gray}{/* \textit{Task prompt} */} \\
We are going over segments of a story sequentially to gradually update one comprehensive summary of the character \{\}.\\ Write a summary for the excerpt provided above, make sure to include vital information related to gender, skills, talents,\\ objectives, background, relationships, key events, and personality of this character. You must briefly introduce characters,\\ places, and other major elements if they are being mentioned for the first time in the summary. The story may feature\\ non-linear narratives, flashbacks, switches between alternate worlds or viewpoints, etc. Therefore, you should organize\\ the summary so it presents a consistent and chronological narrative. Despite this step-by-step process of updating the\\ summary, you need to create a summary that seems as though it is written in one go. The summary must be within \{\} \\words and could include multiple paragraphs.\\ 
\color{gray}{/* \textit{Output Format} */} \\
Output your summary into four specific sections, ...\\
\\
Summary:}\\ 
    \midrule
 \rowcolor[gray]{0.95}\multicolumn{1}{c}{\textbf{II: Update}} \\
 \makecell[l]{
    \color{gray}{/* \textit{Data} */}\\
Below is a segment from a story:\\
\\
- - -\\
\{\}\\
- - -\\
\\
Below is a summary of the character \{\} of the story up until this point:\\
\\
- - -\\
\{\}\\
- - -\\
    \color{gray}{/* \textit{Task prompt} */} \\
We are going over segments of a story sequentially to gradually update one comprehensive summary of the character \{\}. \\You are required to update the summary to incorporate any new vital information in the current excerpt. This information\\ may relate to gender, skills, talents, objectives, background, relationships, key events, and personality of this character.\\ You must briefly introduce characters, places, and other major elements if they are being mentioned for the first time in\\ the summary. The story may feature non-linear narratives, flashbacks, switches between alternate worlds or viewpoints,\\ etc. Therefore, you should organize the summary so it presents a consistent and chronological narrative. Despite this\\ step-by-step process of updating the summary, you need to create a summary that seems as though it is written in one go.\\ The updated summary must be within \{\} words and could include multiple paragraphs.\\ 
\color{gray}{/* \textit{Output Format} */} \\
Output your summary into four specific sections, ...\\
\\
Updated summary:}\\ 
     \midrule
    \rowcolor[gray]{0.95}\multicolumn{1}{c}{\textbf{III: Compress}} \\
    \makecell[l]{
    \color{gray}{/* \textit{Data} */}\\
Below is a segment from a story:\\
\\
- - -\\
\{\}\\
- - -\\
    \color{gray}{/* \textit{Task prompt} */} \\
Currently, this summary contains \{\} words. Your task is to condense it to less than \{\} words. The condensed summary\\ should remain clear, overarching, and fluid while being brief. Whenever feasible, maintain details about gender, skills,\\ talents, objectives, background, relationships, key events, and personality about this character - but express these elements\\ more succinctly. Make sure to provide a brief introduction to characters, places, and other major components during their\\ first mention in the condensed summary. Remove insignificant details that do not add much to the character portrayal.\\ The story may feature non-linear narratives, flashbacks, switches between alternate worlds or viewpoints, etc. Therefore,\\ you should organize the summary so it presents a consistent and chronological narrative.\\ 
\color{gray}{/* \textit{Output Format} */} \\
Output your summary into four specific sections, ...\\
\\
Condensed summary (to be within \{\} words):}\\ 
    \bottomrule
    \end{tabular}
  \caption{Prompt templates for incremental updating.}
  \label{tab:prompt_for_incremental}
\end{table*}
\begin{table*}[t]
  \centering
  \small
    \begin{tabular}{l}
    \toprule
    \rowcolor[gray]{0.95}\multicolumn{1}{c}{\textbf{I: Init}} \\
 \makecell[l]{
    \color{gray}{/* \textit{Data} */}\\
Below is a part of a story:\\
\\
- - -\\
\{\}\\
- - -\\
    \color{gray}{/* \textit{Task prompt} */} \\
We are creating one comprehensive summary for the character \{\} by recursively merging summaries of its chunks. Now,\\ write a summary for the excerpt provided above, make sure to include vital information related to gender, skills, talents,\\ objectives, background, relationships, key events, and personality of this character. You must briefly introduce characters,\\ places, and other major elements if they are being mentioned for the first time in the summary. The story may feature non-\\linear narratives, flashbacks, switches between alternate worlds or viewpoints, etc. Therefore, you should organize the\\ summary so it presents a consistent and chronological narrative. Despite this recursive merging process, you need to create\\ a summary that seems as though it is written in one go. The summary must be within \{\} words and could include multiple\\ paragraphs.\\ 
\color{gray}{/* \textit{Output Format} */} \\
Output your summary into four specific sections, ...\\
\\
Summary:}\\ 
    \midrule
 \rowcolor[gray]{0.95}\multicolumn{1}{c}{\textbf{II: Merge}} \\
 \makecell[l]{
    \color{gray}{/* \textit{Data} */}\\
Below are several summaries of the character \{\} from consecutive parts of a story:\\
\\
- - -\\
\{\}\\
- - -\\
    \color{gray}{/* \textit{Task prompt} */} \\
We are creating one comprehensive summary for the character \{\} by recursively merging summaries of its chunks. Now,\\ merge the given summaries into one single summary, make sure to include vital information related to gender, skills, talents,\\ objectives, background, relationships, key events, and personality of this character. You must briefly introduce characters,\\ places, and other major elements if they are being mentioned for the first time in the summary. The story may feature non-\\linear narratives, flashbacks, switches between alternate worlds or viewpoints, etc. Therefore, you should organize the\\ summary so it presents a consistent and chronological narrative. Despite this recursive merging process, you need to create\\ a summary that seems as though it is written in one go. The summary must be within \{\} words and could include multiple\\ paragraphs.\\ 
\color{gray}{/* \textit{Output Format} */} \\
Output your summary into four specific sections, ...\\
\\
Summary:}\\ 
     \midrule
    \rowcolor[gray]{0.95}\multicolumn{1}{c}{\textbf{III: Merge Context}} \\
    \makecell[l]{
    \color{gray}{/* \textit{Data} */}\\
Below is a summary of the context about the character \{\} preceding some parts of a story:\\
\\
- - -\\
\{\}\\
- - -\\
\\
Below are several summaries of the character \{\} from consecutive parts of the story:\\
\\
- - -\\
\{\}\\
- - -\\
    \color{gray}{/* \textit{Task prompt} */} \\
We are creating one comprehensive summary of the character \{\} by recursively merging summaries of its chunks. Now,\\ merge the preceding context and the summaries into one single summary, make sure to include vital information related to\\ gender, skills, talents, objectives, background, relationships, key events, and personality of this character. You must briefly\\ introduce characters, places, and other major elements if they are being mentioned for the first time in the summary. The\\ story may feature non-linear narratives, flashbacks, switches between alternate worlds or viewpoints, etc. Therefore, you\\ should organize the summary so it presents a consistent and chronological narrative. Despite this recursive merging process,\\ you need to create a summary that seems as though it is written in one go. The summary must be within \{\} words and could\\ include multiple paragraphs.\\ 
\color{gray}{/* \textit{Output Format} */} \\
Output your summary into four specific sections, ...\\
\\
Summary:}\\ 
    \bottomrule
    \end{tabular}
  \caption{Prompt templates for hierarchical merging.}
  \label{tab:prompt_for_hierarchical}
\end{table*}
\begin{table*}[t]
  \centering
  \small
    \begin{tabular}{l}
    \toprule
    \rowcolor[gray]{0.95}\multicolumn{1}{c}{\textbf{I: Consistency Score}} \\
    \makecell[l]{
    \color{gray}{/* \textit{Task prompt} */}\\
    You are a character extraction performance comparison assistant. You will be given the golden information about character\\ \{\}'s \{dimension\} in a novel. You will then be given the summarized information about character \{\} extracted by a model\\ from the origin novel.\\
    Your task is to rate the summarized information on one metric.\\
    Please make sure you read and understand these instructions carefully.\\
    \\
    Evaluation Criteria:\\
    Consistency (1-5) - the factual alignment between the golden and the summarized information. A score of 1 indicates\\ significant discrepancies, while a score of 5 signifies a high level of factual consistency.\\
    \\
    Evaluation Steps:\\
    1. Read the golden information carefully and identify the main facts and details it presents.\\
    2. Read the summarized information and compare it to the golden information. Check if the summary contains any factual\\ errors or lacks necessary foundational facts. If the summarized one includes information not mentioned in the golden\\ information, please ignore it, as the summary is extracted from the original book and may contain more extraneous \\information.\\
    3. Assign a score for consistency based on the Evaluation Criteria and explain the reason. Your output should be structured\\ as the following schema: \{\{``score'': int // A score range from 1 to 5, ``reason'': string // The reason of evaluation result\}\}\\
    \color{gray}{/* \textit{Data} */}\\
    Golden information: \\
    \{\}\\
    Summarized information: \\
    \{\}\\
    \color{gray}{/* \textit{Output Format} */}\\
    Evaluation Form (Please output the result in JSON format. Do not output anything except for the evaluation result. All\\ output must be in JSON format and follow the schema specified above.):\\
    - Consistency:\\
    \color[rgb]{0,0.39,0}\textit{\{}\\ 
    \color[rgb]{0,0.39,0}\textit{\quad``score'': 3,}\\ 
    \color[rgb]{0,0.39,0}\textit{\quad``reason'': ``The summarized information is partially consistent with the golden information, ...''}\\ 
    \color[rgb]{0,0.39,0}\textit{\}}} \\
    \midrule
    \end{tabular}
  \caption{Prompt templates for factual consistency examination. Generated texts by GPT-4 are {\color[rgb]{0,0.39,0}\textit{highlighted}}.
  }
  \label{tab:prompt_for_fce}
\end{table*}
\begin{table*}[t]
  \centering
  \small
    \begin{tabular}{l}
    \toprule
    \makecell[l]{
    \rowcolor[gray]{0.95}\multicolumn{1}{c}{\textbf{I: Normal}} \\
    \color{gray}{/* \textit{Task prompt} */}\\
    You are a helpful assistant proficient in analyzing the motivation for the character's decision in novels. You will be given the\\ profile about character \{\} in a novel. Your task is to choose the most accurate primary motivation for the character's decision\\ according to the character's profile. You also need to provide reasons, the reasons should be related to the character's basic\\ attributes, experiences, relationships, or personality, of this character.\\
    Your output should be structured as the following schema: \\
    \{\{``Choice'': str // ``A''/``B''/``C''/``D'', ``Reason'': string // The reason of the choice\}\}\\
    \color{gray}{/* \textit{Data} */} \\
    Character Profile:\\
    name: \{\}\\
    Summary of this character: \{\}\\
    \\
    Question:\\
    \{\}\\
    \color{gray}{/* \textit{Output Format} */} \\
    Output (All output must be in JSON format and follow the schema specified above.):\\
    \color[rgb]{0,0.39,0}\textit{\{}\\ 
    \color[rgb]{0,0.39,0}\textit{\quad``Choice'': ``A'',}\\ 
    \color[rgb]{0,0.39,0}\textit{\quad``Reason'': ``Margot's primary motivation for ...''}\\ 
    \color[rgb]{0,0.39,0}\textit{\}}\\
    \rowcolor[gray]{0.95}\multicolumn{1}{c}{\textbf{II: Ablate All Dimensions}} \\
    \color{gray}{/* \textit{Task prompt} */}\\
    You are a helpful assistant proficient in analyzing the motivation for the character's decision in novels. Your task is to choose\\ the most accurate primary motivation for the character's decision according to the character's profile. Since you are not given\\ the character analysis, you are supposed to choose the most reasonable motivation based on the provided information in the\\ question.\\
    Your output should be structured as the following schema: \\
    \{\{``Choice'': str // ``A''/``B''/``C''/``D'', ``Reason'': string // The reason of the choice\}\}\\
    \color{gray}{/* \textit{Data} */} \\
    Character Profile:\\
    name: \{\}\\
    \\
    Question:\\
    \{\}\\
    \color{gray}{/* \textit{Output Format} */} \\
    Output (All output must be in JSON format and follow the schema specified above.):\\
    \color[rgb]{0,0.39,0}\textit{\{}\\ 
    \color[rgb]{0,0.39,0}\textit{\quad``Choice'': ``A'',}\\ 
    \color[rgb]{0,0.39,0}\textit{\quad``Reason'': ``Given the lack of specific information about Margot, ...''}\\ 
    \color[rgb]{0,0.39,0}\textit{\}}}\\
    \bottomrule
    \end{tabular}
  \caption{Prompt templates for motivation recognition. Generated texts by GPT-4 are {\color[rgb]{0,0.39,0}\textit{highlighted}}.}
  \label{tab:prompt_for_mr}
\end{table*}\label{sec:appendix}
\end{document}